\def\year{2021}\relax
\documentclass[letterpaper]{article} 
\usepackage{aaai21}  
\usepackage{times}  
\usepackage{helvet} 
\usepackage{courier}  
\usepackage[hyphens]{url}  
\usepackage{graphicx} 
\usepackage{natbib}  
\usepackage{caption} 
\usepackage{subcaption}
\usepackage[utf8]{inputenc} 
\usepackage[T1]{fontenc}    
\usepackage{url}            
\usepackage{booktabs}       
\usepackage{amsfonts}       
\usepackage{nicefrac}       
\usepackage{microtype}      
\usepackage{xcolor}
\usepackage{amsmath}
\usepackage{mathtools}
\usepackage{amsthm}
\usepackage{graphicx}
\usepackage{caption}
\usepackage{lineno}
\usepackage{comment}
\usepackage{cite}
\usepackage{amsmath,amssymb,amsfonts}
\usepackage{comment}
\usepackage{float}
\usepackage{multirow}
\usepackage{makecell}
\usepackage{caption}
\usepackage{subcaption}
\usepackage{graphicx}
\usepackage{textcomp}
\usepackage{xcolor}
\usepackage{setspace}
\usepackage{algorithmic}
\usepackage[ruled,linesnumbered]{algorithm2e}
\DeclarePairedDelimiterX\set[1]\lbrace\rbrace{#1}

\newcommand*{\email}[1]{\texttt{#1}}
\newcommand*\Mname{\textcolor{black}{UAG}}

\title{Uncertainty-aware Attention Graph Neural Network for Defending Adversarial Attacks}

\author{
    Boyuan Feng, Yuke Wang, Zheng Wang, and Yufei Ding \\
}

\affiliations{
    University of California, Santa Barbara, USA \\
    \email{boyuan@cs.ucsb.edu, yuke\_wang@cs.ucsb.edu} \\ 
    \email{zhengwang0122@gmail.com, yufeiding@cs.ucsb.edu}\\
}

\begin{document}

\maketitle

\begin{abstract}
With the increasing popularity of graph-based learning, graph neural networks (GNNs) emerge as the essential tool for gaining insights from graphs.
However, unlike the conventional CNNs that have been extensively explored and exhaustively tested, people are still worrying about the GNNs' robustness under the critical settings, such as financial services.
The main reason is that existing GNNs usually serve as a black-box in predicting and do not provide the uncertainty on the predictions.
On the other side, the recent advancement of Bayesian deep learning on CNNs has demonstrated its success of quantifying and explaining such uncertainties to fortify CNN models.
Motivated by these observations, we propose UAG, the first systematic solution to defend adversarial attacks on GNNs through identifying and exploiting hierarchical uncertainties in GNNs.
UAG develops a Bayesian Uncertainty Technique (BUT) to explicitly capture uncertainties in GNNs and further employs an Uncertainty-aware Attention Technique (UAT) to defend adversarial attacks on GNNs. Intensive experiments show that our proposed defense approach outperforms the state-of-the-art solutions by a significant margin.
\end{abstract}

\section{Introduction} \label{sec:intro}
As the emerging trend of extending deep learning from Euclidean data (\textit{e.g.}, images) to non-Euclidean data (\textit{e.g.}, graphs), graph neural network (GNN)~\cite{GINConv,GCNConv,GATConv} wins lots of attentions from both research and industrial domains. 
Compared with the conventional graph-learning approaches (\textit{e.g.}, random walk~\cite{grover2016node2vec, deepWalk}, and graph Laplacians~\cite{luo2011cauchy, luo2009non, cheng2018deep}), GNNs excel at both computation efficiency and runtime performance for various tasks, such as the node classification~\cite{kaspar2010graph, gibert2012graph, duran2017learning} and link prediction~\cite{chen2005link, kunegis2009learning, tylenda2009towards}.
Despite the stunning success, people still concern about the robustness of GNNs, especially in some safety-critical domains (\textit{e.g.}, financial services and medicinal chemistry).
Existing work \cite{zugner2019adversarial, xu2019topology, DICE} has shown that GNNs are sensitive to small perturbations on the topology and the node features, which motivates our work for defending adversarial attacks.

The most recent work, RGCN~\cite{zhu2019robust}, improves GNN robustness with a simple strategy that replaces deterministic GNN features with a Gaussian distribution and measures the variance in the intermediate feature vectors.
However, it assumes fixed GNN weights without quantifying the uncertainty from GNN models and does not consider the uncertainty from the graph topology, leading to unsatisfactory accuracy under severe attacks.
We believe the key to improve the GNN robustness is to develop a powerful technique to quantify and exploit uncertainties from various sources to absorb the effect of adversarial attacks.



In this paper, we focus on exploring the benefits of explicitly quantifying GNN uncertainty to defend GNNs against adversarial attacks. 
And our further investigation shows that there are two major types of uncertainties in GNNs -- the \textit{model} uncertainty and the \textit{data} uncertainty.
The former refers to the uncertainty in model parameters to tell whether the selected parameters can best suit the distribution of the collected data.
The latter refers to the uncertainty in the noisy data collection, coming from either the noises in the data collection process or the adversarial attacks.
However, exploring these uncertainties is non-trivial since there are several challenges to overcome: 
\begin{enumerate} 
\item \textbf{Uncertainty Measurement:} How to explicitly measure the uncertainty of GNNs?
\item \textbf{Robustness:} How to effectively incorporate the measured uncertainty into existing
GNNs for defending adversarial attacks?
\end{enumerate}

To tackle these challenges, we propose the first Bayesian-based uncertainty guided approach to defend the GNN effectively. 
First, we develop a \textit{Bayesian Uncertainty Technique} (BUT) based on the powerful Bayesian framework to capture these uncertainties from different sources.
Intuitively, we measure the uncertainty value for individual nodes where a higher uncertainty usually indicates a lower prediction accuracy.
Then, we design an \textit{Uncertainty-aware Attention Technique} (UAT) to dynamically adjust the impact of one node towards its neighboring nodes according to its uncertainty.
In particular, for nodes with high uncertainty that may have been attacked, we restrict its feature propagation towards neighboring nodes in order to absorb the attack impact.








In short, we summerize our contributions as follows:
\begin{itemize}
    \item We identify two types of uncertainties in GNNs (\textit{i.e.}, model uncertainty and data uncertainty) and propose a Bayesian Uncertainty Technique (BUT) to explicitly capture both types of uncertainties.  
    \item We introduce an Uncertainty-aware Attention Technique (UAT) to defend the adversarial attack by assigning less impact (weights) on nodes with high uncertainty, thus, mitigating their impact on the final prediction.
    \item Rigorous experiments and studies across various datasets on mainstream GNNs show that our proposed defense approach outperforms the state-of-the-art RGCN by a significant margin.
\end{itemize}
\section{Related Work} \label{sec:related_work}

\textbf{Graph Neural Network}
Graph Neural Networks (GNNs) are now becoming a major way of gaining insights from the graph structures. It generally includes several graph convolutional layers, each of which consists of a neighbor aggregation and a node update step. 
The most common graph convolutional layer \cite{GCNConv} computes the embedding for node $v$ at layer $k+1$ based on node embedding at layer $k$, where $k \geq 0$.
\begin{equation} \label{eq: GNN} \small
    h_v^{(k+1)} = \sigma(\sum_{u\in \Bar{N}(v)} \frac{1}{c_u c_v}h_u^{(k)} \cdot W^{(k)})
\end{equation}
As shown in Equation~\ref{eq: GNN}, $h_{v}^{(k)}$ is the embedding vector for node $v$ at layer $k$, $W^{(k)}$ is the GNN weight at layer $k$, and $\Bar{N}(v) = N(v) \cup v$ is the set of node $v$ and its neighboring nodes.
$c_u$ and $c_v$ are fixed values determined by the degree of node $u$ and $v$ and will be omitted in following sections for notation simplicity.
Intuitively, the graph convolution layer aggregates information across nodes by averaging features in nearby nodes.
More advanced GNNs utilize different aggregation methods.
For example, GAT \cite{GATConv} aggregates node features with weighted average based on the cosine similarity between node features.


\vspace{2pt}
\noindent \textbf{Graph Adversarial Attacks and Defense}
Existing works have explored the robustness of the GNNs in two opposite but closely related directions, GNN attacks, and GNN defense.
On the attack side, existing GNN attacks can be broadly classified into two major categories, \textit{poisoning}~\cite{zugner2018adversarial,zugner2019adversarial} and \textit{evasion}~\cite{dai2018adversarial}, depending on the time they happen.
The former (poisoning attack) happens during the training time of the GNNs through modifying training data and the latter (evasion attack) takes place during the GNN inference time by changing test data samples. Our work is orthogonal and complementary to these existing GNN attack research, since 1) our goal is to minimize the impact of these GNN attacks by incorporating model and data uncertainties during the GNN computation; 2) our defense-oriented research may potentially motivate more diverse adversarial attacks tailored for GNNs.

On the defense side, RGCN~\cite{zhu2019robust} proposes a novel model to make GCN immune from adversarial attacks by leveraging Gaussian distributions to reduce the impact of GNN attacks. 
Different from RGCN, our \Mname~is the first work to identify and quantify how adversarial attacks affect GNN's performance -- model and data uncertainties that take both model (weight) and data (topology and embedding features) into consideration.
And we further exploit such uncertainty information by incorporating a novel technique (\textit{e.g.}, BUT and UAT) to facilitate the defense.

\vspace{2pt}
\noindent \textbf{Bayesian Neural Network and Uncertainty}
Many research efforts have been made towards developing Bayesian Neural Network to measure uncertainty in computer vision \cite{kendall2017uncertainties, dropoutBayesian, dropoutRNN, BMVC2017_57}, natural language processing \cite{xiao2019quantifying}, and time series analysis \cite{uberUncertainty}.
These works usually focus on convolutional neural networks and use Bayesian Neural Network as a regularization technique.
Some recent contributions \cite{BGCN, hasanzadeh2020bayesian} extend the Bayesian Neural Network to the graph domain as a stochastic regularization technique.
These works aim to solve the over-smoothing problem in GNNs and do not explicitly quantify the uncertainty.
To the best of our knowledge, we are the first to explicitly quantify the uncertainty in the graph domain and exploit the uncertainty to defend adversarial attacks.


\section{Methodology} \label{sec:methodology}
\begin{figure}[t]
    \centering
    \includegraphics[width=0.95\columnwidth]{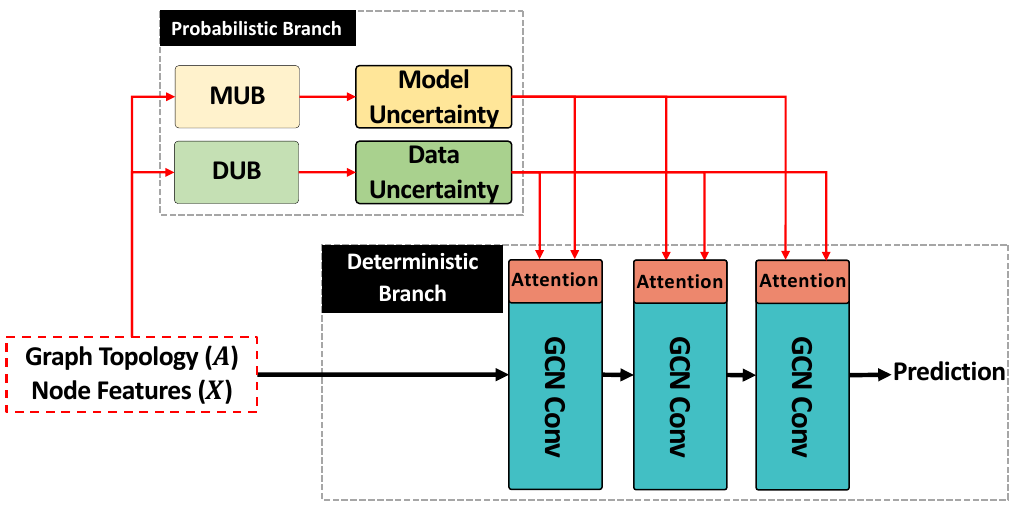}
    \caption{Overview of \Mname.} 
    \vspace{-15pt}
    \label{fig: Overview}
\end{figure}

\begin{figure*}[t]
    \centering
    \includegraphics[width=0.75\textwidth]{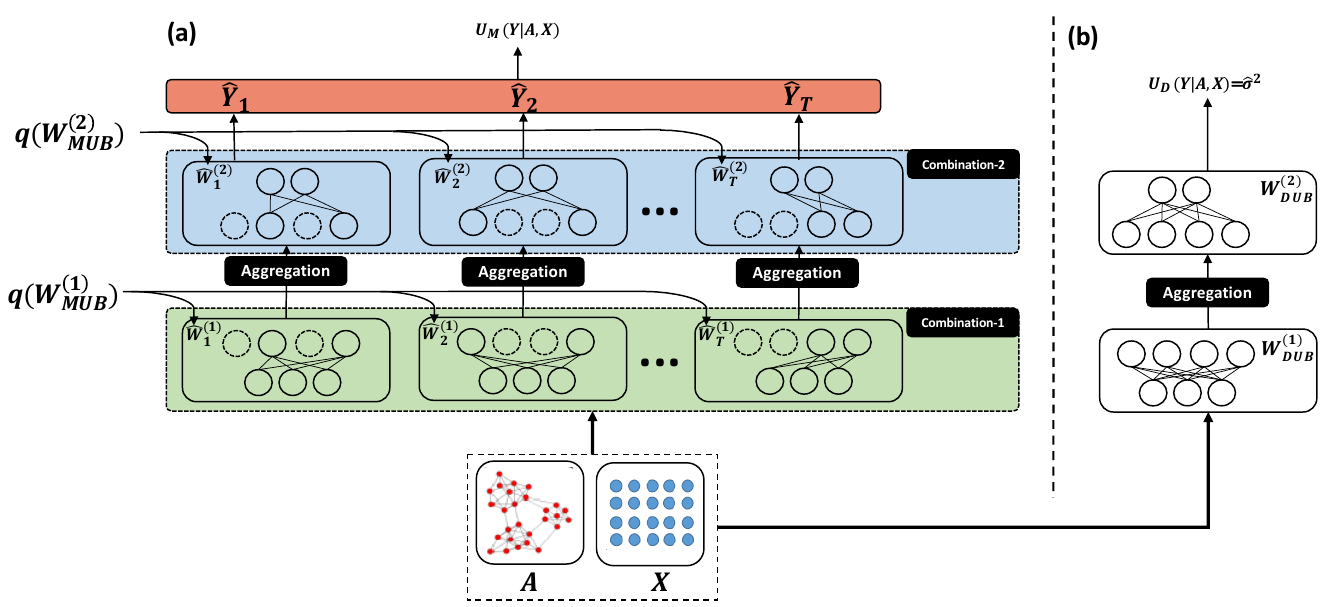}
    \vspace{-5pt}
    \caption{Bayesian Uncertainty Technique Overview. (a) Model Uncertainty Branch (MUB); (b) Data Uncertainty Branch (DUB)}
    \label{fig:illustrationBUT}
    \vspace{-10pt}
\end{figure*}

The overview of UAG is presented in Figure \ref{fig: Overview}.
UAG takes two inputs -- the adjacency matrix $A \in \mathcal{R}^{N\times N}$ and the node features $X \in \mathcal{R}^{N \times D}$, where $N$ is the number of nodes and $D$ is the dimension of node features.
There are two main branches in UAG -- the probabilistic branch (including the Model Uncertainty Branch (MUB) and the Data Uncertainty Branch (DUB)) and the deterministic branch, where the architecture and weights are different across branches.
Given the graph data $(A,X)$, the probabilistic branch measures the node-wise uncertainty $U = [U_M, U_D]\in \mathcal{R}^{N \times 2}$ from the GNN model weights and the graph data.
Here, the probabilistic branch adopts a novel Bayesian Uncertainty Technique (BUT) to measure the uncertainty for each node due to the adversarial attacks.
The deterministic branch takes the measured node-wise uncertainty $U$ and the graph data $(A,X)$ to generate the node classification results $Y \in \mathcal{R}^{N}$.
It contains an Uncertainty-aware Attention Technique (UAT) to adaptively adjust the edge attention during the inference to defend against adversarial attacks.

\subsection{Bayesian Uncertainty Technique}

Bayesian uncertainty technique measures two sources of uncertainties -- the model uncertainty $U_M \in \mathcal{R}^{N}$ and the data uncertainty $U_D \in \mathcal{R}^N$.
Formally, following the law of total variance, the uncertainty in the prediction $Y$ is
\begin{equation} \small
\begin{split}
    Var(Y) & = Var(E[Y|A,X]) + E[Var(Y|A,X)] \\
           & = U_M(Y|A,X) + U_D(Y|A,X)
\end{split}
\end{equation}
Here, we use a model uncertainty branch (MUB) and a data uncertainty branch (DUB) to access $U_M(Y|A,X)$ and $U_D(Y|A,X)$, respectively.



\vspace{4pt}
\noindent \textbf{Model Uncertainty.}
The model uncertainty $U_M$ measures the uncertainty in the mapping process $E[Y|A,X]$ due to model weight selection.
We use a 2-layer GCN to quantify the model uncertainty.
Instead of using fixed weights, we utilize a probability distribution to describe the uncertainty from the model weights, as illustrated in Figure \ref{fig:illustrationBUT}(a).
Formally, given the graph data $(A,X)$ and the partial label $Y \in \mathcal{R}^{N_L}$ with $N_L$ as the number of labeled nodes, we first train the weight posterior distribution $p(W|A,X)$.
Then, we conduct the prediction mapping procedure as 
\begin{equation}\small
    p(Y|X,A) = \int_W p(Y|W,A,X) \; p(W|A,X) dW
\end{equation}
Since the exact inference is intractable, we adopt the MC dropout variational inference \cite{dropoutBayesian} method by multiplying a deterministic model weight $W_{MUB}$ with a random variable $B$ following the Bernoulli distribution.
This provides $q(W)$ as an approximation to the true posterior $p(W|A,X)$.
In particular, the model weights $W$ follows
\begin{equation}\small
\begin{split}
    q(W) & \sim B \odot W_{MUB} \\
    P(B) & \sim Bernoulli(p)
\end{split}
\end{equation}
where $\odot$ is the Hadamard product, and $p$ is a hyperparameter (=0.8 by default in our evaluation).
During training, we can train the weight by minimizing the cross-entropy loss
\begin{equation} \small
    L_{model} = -\frac{1}{T} \sum_{t=1}^T log \; p(\hat{Y}_t|\hat{W}_t, A, X) + \frac{1-p}{2T} ||W_{MUB}||^2
\end{equation}
where $\hat{W}_t$ is sampled from $q(W)$, $\hat{Y}_t$ is the prediction under sampled weight $\hat{W}_t$, and $T$ is the number of samples during the MC dropout variational inference.
During inference, we perform the Monte Carlo integration:
\begin{equation} \small \label{eq:modelUncertaintyI}
    E(Y|A,X) = \frac{1}{T} \sum_{t=1}^T \hat{Y}_t
\end{equation}
where $\hat{Y}_t \in \mathcal{R}^L$ is the prediction after the softmax layer, and $L$ is the number of classes in the graph data.

Here, we observe that applying Bernoulli distribution at different granularities leads to different probabilistic interpretation.
To provide a comprehensive measurement on the model uncertainty, we apply dropout independently for individual GNN layers, channels, nodes, and edges
\begin{equation} \small \label{eq:sample}
    W_{uv}^{(k)} = B^{(k)}_{uv} \odot B^{(k)}_u \odot W^{(k)}
\end{equation}
where $W^{(k)} \in \mathcal{R}^{f_k \times f_{k+1}}$ is the GNN weights at the $k^{th}$ layer, $B^{(k)}_{uv} \in \mathcal{R}$ determines the dropout on the edge-level, $B^{(k)}_u \in \{0,1\}^{f_k}$ drops the weight at the channel level, and $f_k$ is the number of feature channels at layer $k$.
From the perspective of individual nodes, we have
\begin{equation} \small
\begin{split}
    h_v^{(k+1)} &= \sigma(\sum_{u\in \Bar{N}(v)} h_u^{(k)} \cdot W_{uv}^{(k)}) \\
                &= \sigma(\sum_{u\in \Bar{N}(v)} h_u^{(k)} \cdot (B^{(k)}_{uv} \odot B^{(k)}_u \odot W^{(k)}))
\end{split}
\end{equation}
where $\Bar{N}(v) = N(v) \cup v$.
Noting that this dropout also provides a Bayesian view of dropping edges or nodes when either $B_{uv}^{(k)} = 0$ or $B_u^{(k)} = 0$.

Given the Bayesian framework on GNNs, we can measure the model uncertainty as the variance in predictions
\begin{equation} \small \label{eq:ModelUncertainty}
\begin{split}
    U_M(Y|A,X) & = Var(Y|A,X) \\
               & = E(Y^2|A,X) - [E(Y|A,X)]^2 \\
               & = \frac{1}{T} \sum_{t=1}^T \hat{Y}_t^2 - [E(Y|A,X)]^2\\
\end{split}
\end{equation}

Here, we additionally apply a reduce operation to transform the $L$-dimension vector $U_M$ to a scalar value as the model uncertainty.
\begin{figure} \small
    \centering
    \includegraphics[width=0.75\linewidth]{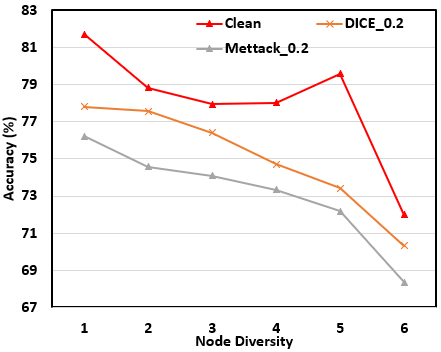}
    \vspace{-5pt}
    \caption{Relationship between accuracy and node diversity. $DICE\_0.2$ and $Mettack\_0.2$ indicates perturbing 20\% edges with DICE and Mettack, respectively.}
    \label{fig:observation}
    \vspace{-10pt}
\end{figure}

\vspace{3pt}
\noindent \textbf{Data Uncertainty}
The data uncertainty $U_D$ measures the prediction noise intrinsic to the data inputs.
There are two standard approaches to identify the data uncertainty on each node.
One approach utilizes the maximum predicted probability to measure confidence in the prediction.
However, this confidence comes as a side effect of the model training and lacks sophisticated probabilistic interpretation.
The other approach from the CNN domain predicts the uncertainty value based on the image inputs.
However, naively borrowing this approach into the GNN domain focuses only on the node features and fails to exploit the important topology information in the graph data.


Instead, we aim to capture the data uncertainty that is intrinsic to the graph topology in terms of node diversity $Div_{node}^{(k)}$, defined as the number of different labels in the node's k-hop neighbors.
Our key observation is that adversarial attacks on graph data usually increase the node diversity and add edge connections between nodes with different labels.
For example, DICE attack (delete edges internally, connect externally)~\cite{DICE} exploits the node label information to increase node diversity by deleting edges between nodes with the same label and adding edges between nodes with different labels.
Figure \ref{fig:observation} shows the accuracy among nodes that have node diversity larger than various thresholds.
We observe that the accuracy usually decreases significantly as the node diversity increases, which holds for both the clean graph data and the attacked graph data from various attacking algorithms.


To this end, we explicitly measure the data uncertainty by treating the prediction as a Gaussian distribution and setting the variance to be the node diversity, as illustrated in Figure \ref{fig:illustrationBUT}(b).
Formally, we have
\begin{equation} \small 
    Y \sim N(\hat{\mu}(A,X), \hat{\sigma}^2(A,X))
\end{equation}
where $\hat{\mu}$ and $\hat{\sigma}^2$ are the predicted label and node diversity, respectively.
Here, we parameterize the $\hat{\mu}$ and $\hat{\sigma}^2$ with the adjacency matrix A and the node feature X and use a 2-layer GCN to predict their values.
During inference, we will use the $\hat{\sigma}^2(A,X)$ as the data uncertainty 
\begin{equation} \small \label{eq:dataUncertainty}
    U_D(Y|A,X) = \hat{\sigma}^2(A,X)
\end{equation}


To train the data uncertainty, we have two losses on the labeled nodes and unlabeled nodes, respectively.
On the labeled nodes, we focus on the ground truth labels and have a KL-divergence that requires the predicted distribution to match with the ground truth distribution
\begin{equation} \small \label{eq:loss1}
    L_1 = KL(N(\hat{\mu}(A,X), \hat{\sigma}^2(A,X)) | N(Y, \sigma^2))
\end{equation}
where the $Y$ comes from the ground truth label and $\sigma^2$ measures the node diversity $Div_{node}^{(k)}$ in the graph data.
When computing the node diversity, we utilize only labeled node in the clean graph data.
In particular, for a given node, we first collect all labeled 2-hop neighboring nodes and then count the number of distinct labels.
On the unlabeled nodes, similar to the unsupervised learning \cite{bojchevski2018deep} on graph data, we focus on the graph topology and adopt an energy-based unsupervised loss 
\begin{equation} \small \label{eq:loss2}
\begin{split} 
    L_2 & = \sum_i \sum_{k<l} \sum_{j_k \in N_{ik}} \sum_{j_k \in N_{il}} (E_{ij_k}^2 + exp^{-E_{ij_l}})    \\
    E_{ij} & = D_{KL}(N(\hat{Y}_j, \hat{\sigma}^2_j)||N(\hat{Y}_i, \hat{\sigma}^2_i))
\end{split}
\end{equation}
Assuming that a node tends to have a similar label with neighboring nodes, this loss implicitly captures the node diversity by forcing higher feature similarities in neighboring nodes.

\begin{figure}[t] \small
    \centering
    \includegraphics[width=0.8\columnwidth]{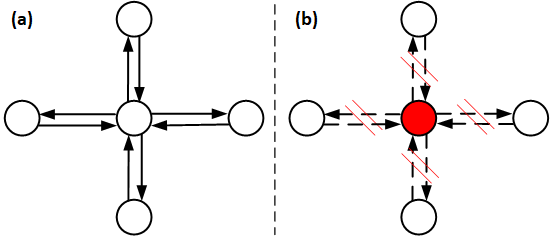}
    \caption{Illustration of UAT. (a) Aggregation on the clean graph; (b) aggregation when the red node is attack.}
    \label{fig:UATIllustration}
    \vspace{-10pt}
\end{figure}

\subsection{Uncertainty-aware Attention Technique}

The uncertainty-aware attention technique (UAT) adaptively adjusts the edge attention during the inference to defend against adversarial attacks, as illustrated in Figure \ref{fig:UATIllustration}.
We equip a 2-layer GCN with our UAT.
On a clean graph (Figure \ref{fig:UATIllustration}a), we adopt edge aggregation similar to existing GNNs that each node aggregates and propagates information across neighboring nodes.
On an attacked graph (\textit{e.g.}, the red node in Figure \ref{fig:UATIllustration}b), UAT adaptively limits the information propagation between the attacked node and other nodes.
While existing works \cite{zugner2018adversarial} have shown that attacking one node in the graph can also lead to the wrong prediction on other nodes, UAT mitigates it by reducing the impact from attacked nodes to remaining nodes.

Formally, given the feature $h_u^{(k+1)}$ for each node $u$ at the $k+1$ layer, we compute an attention $Att_\tau(u)$ for each node $u$ and compute each GNN layer as
\begin{equation} \small \label{eq:UAT}
\begin{split}
   h_v^{(k+1)} & = \sigma(\sum_{u\in \Bar{N}(v)} Att_\tau^{uv}  \cdot h_u^{(k)} \cdot W^{(k)}) \\
   Att_\tau^{uv} & = min(Att_\tau(u), Att_\tau(v))
\end{split}
\end{equation}
where $\tau \in \{M, D\}$ indicates whether we are using model uncertainty or the data uncertainty, each node embedding $h_u^k$ is weighted by an attention.
Here, we use attention from both nodes $u$ and $v$ to decide the attention value on the edge.
Note that the deterministic branch focuses on improving accuracy and utilizes independent weight from the probabilistic branch, which focuses on capturing uncertainties.
We design two attentions to measure the model uncertainty and the data uncertainty, respectively
\begin{equation} \small \label{eq:deterministicInference}
\begin{split}
    Att_{\tau}(u) & = exp(-\zeta \cdot U_{\tau, u}) \\
    \zeta & = \alpha_\tau \cdot exp(-\beta_\tau \cdot Range(U_{\tau}))
\end{split}
\end{equation}
where $U_{\tau,u}$ measures the uncertainty on node $u$, and $\alpha_{\tau} >0$ and $\beta_{\tau}>0$ are two hyperparameters to adjust the impact from uncertainty.
Intuitively, a larger uncertainty $U_{\tau,u}$ on a node $u$ leads to lower weight in the information propagation along with the graph topology.
Here, we additionally utilize a $Range(U_\tau)$ operation to measure the global uncertainty diversity in order to absorb the uniform uncertainty scale change on all nodes under diverse attacks.
In particular, we collect $U_\tau$ for all nodes and measure the $Range(U_\tau)$ as the absolute difference between the first and the third quartiles.
We have investigated several functions to combine the data uncertainty and the model uncertainty, and find out a simple minimal combination can already lead to good performance
\begin{equation} \small
    Att_{Both}(u) = min(Att_M, Att_D)
\end{equation}
Intuitively, we restrict the information propagation from one node when it shows either a high model uncertainty or a high data uncertainty.

\begin{figure*}
    \begin{subfigure}{0.32\textwidth}
        \centering
        \includegraphics[width=\textwidth]{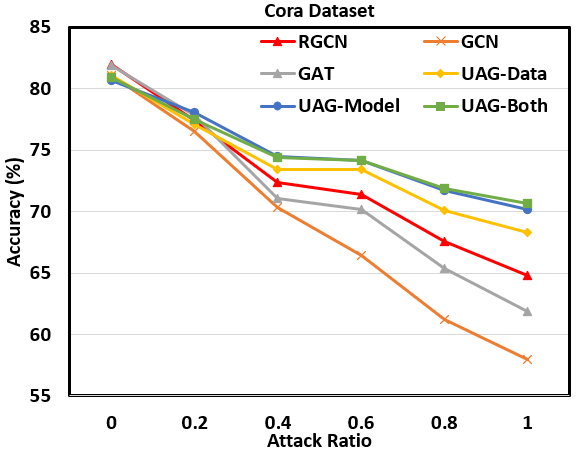}
    \end{subfigure}
    \begin{subfigure}{0.32\textwidth}
        \centering
        \includegraphics[width=\textwidth]{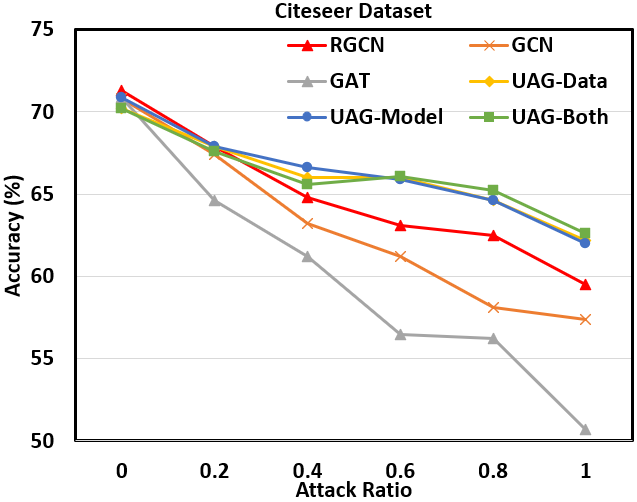}
    \end{subfigure}
    \begin{subfigure}{0.32\textwidth}
        \centering
        \includegraphics[width=\textwidth]{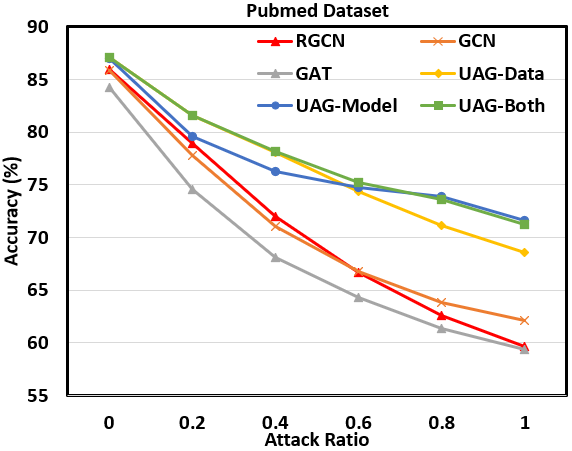}
    \end{subfigure}
    \caption{Results of different methods when adopting \textbf{Random Attack} as the attack method.}
    \label{fig:Random}
\end{figure*}

\begin{figure*}
    \begin{subfigure}{0.32\textwidth}
        \centering
        \includegraphics[width=\textwidth]{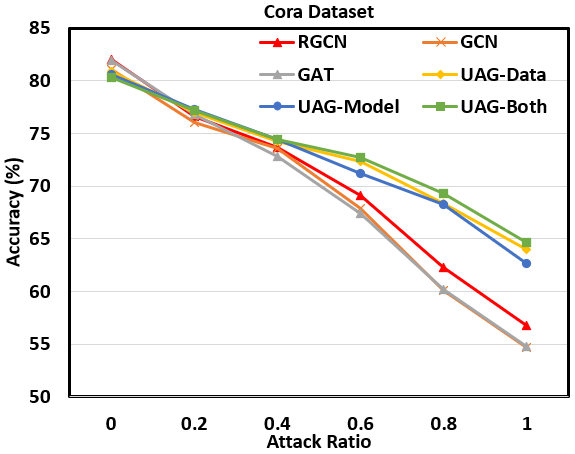}
    \end{subfigure}
    \begin{subfigure}{0.32\textwidth}
        \centering
        \includegraphics[width=\textwidth]{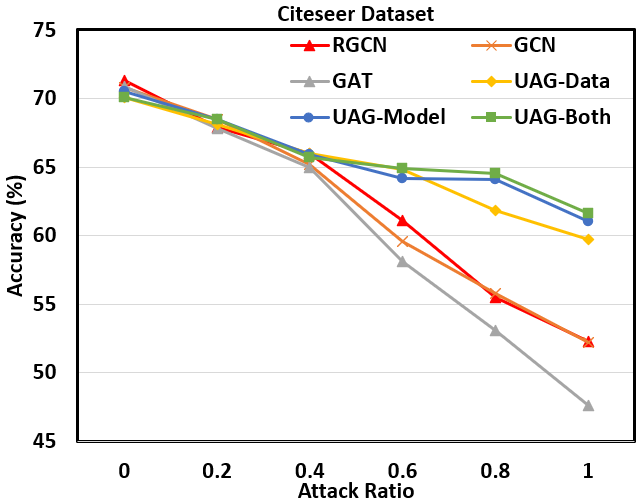}
    \end{subfigure}
    \begin{subfigure}{0.32\textwidth}
        \centering
        \includegraphics[width=\textwidth]{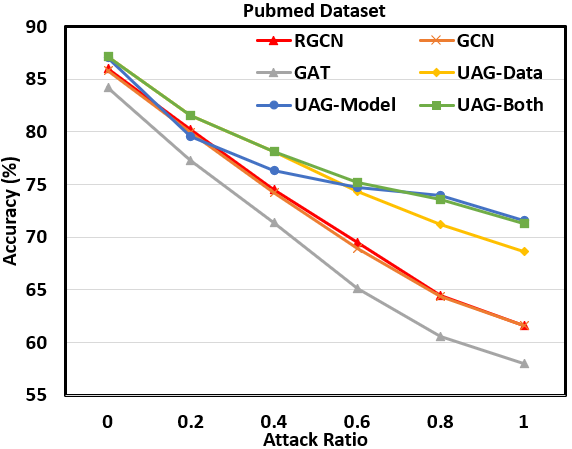}
    \end{subfigure}
    \caption{Results of different methods when adopting \textbf{DICE Attack} as the attack method}
    \label{fig:DICE}
\end{figure*}

\begin{figure*}
    \begin{subfigure}{0.32\textwidth}
        \centering
        \includegraphics[width=\textwidth]{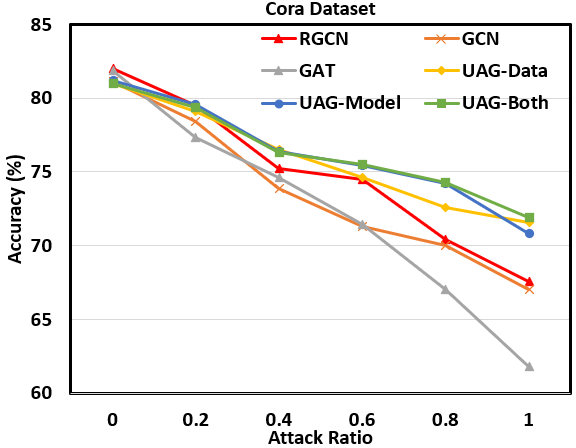}
    \end{subfigure}
    \begin{subfigure}{0.32\textwidth}
        \centering
        \includegraphics[width=\textwidth]{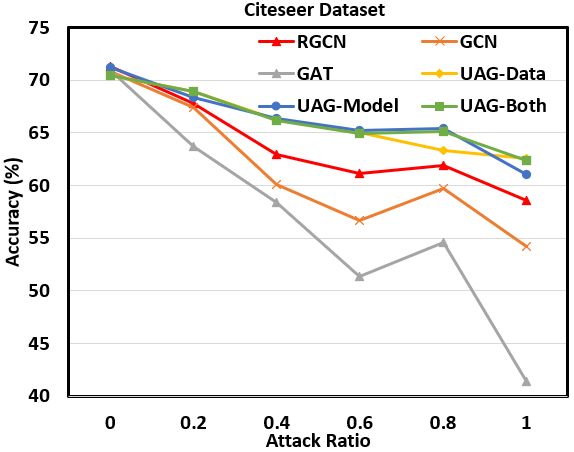}
    \end{subfigure}
    \begin{subfigure}{0.32\textwidth}
        \centering
        \includegraphics[width=\textwidth]{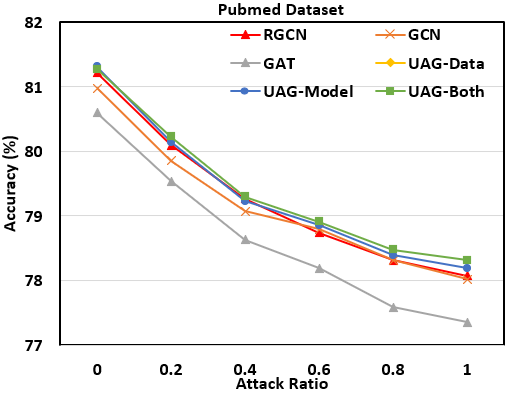}
    \end{subfigure}
    \caption{Results of different methods when adopting \textbf{Mettack} as the attack method}
    \label{fig:Mettack}
\end{figure*}

\begin{figure*}
    \begin{subfigure}{0.32\linewidth}
        \centering
        \includegraphics[width=\textwidth]{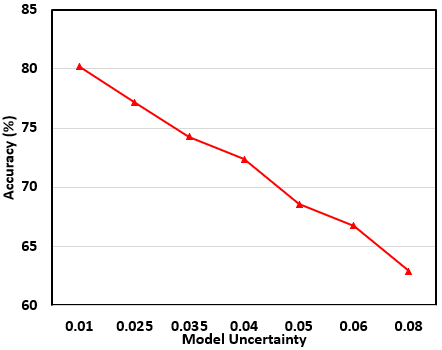}
    \end{subfigure}
    \begin{subfigure}{0.32\linewidth}
        \centering
        \includegraphics[width=\textwidth]{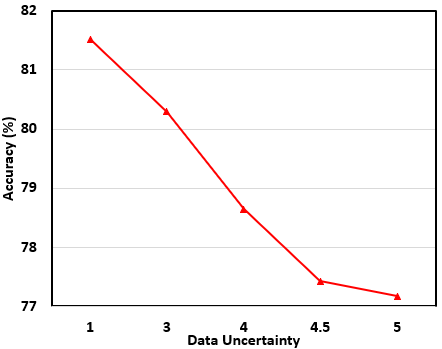}
    \end{subfigure}
    \begin{subfigure}{0.32\textwidth}
        \centering
        \includegraphics[width=\textwidth]{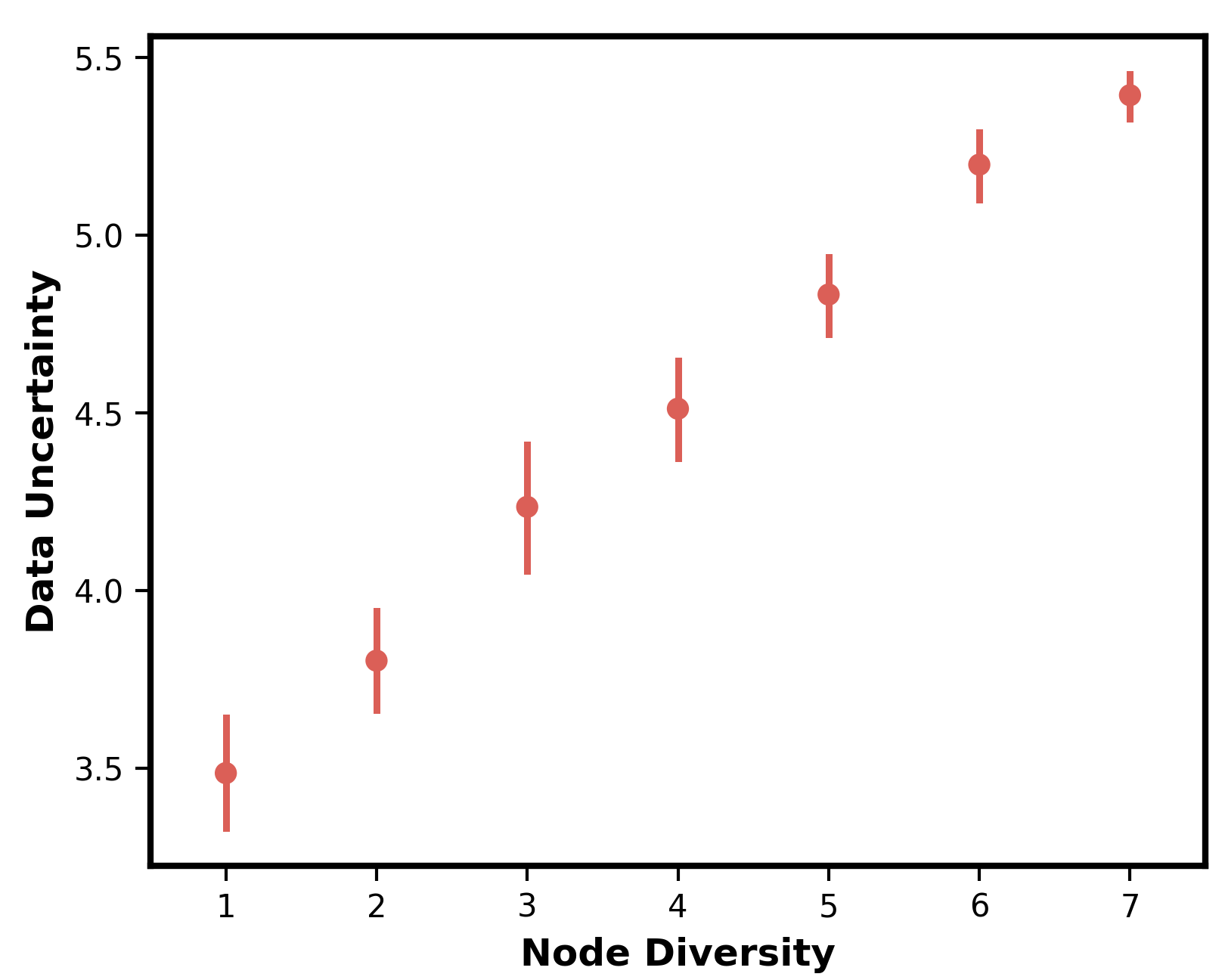}
    \end{subfigure}
    \caption{Relationship between Accuracy and Uncertainty. \textbf{Left:} Model Uncertainty v.s. Accuracy. \textbf{Mid:} Data Uncertainty v.s. Accuracy. \textbf{Right:} Data Uncertainty v.s. True Diversity.}
    \label{fig:RelationshipAccuracy_Uncertainty_Diversity}
\end{figure*}

\begin{figure*}[t]
    \begin{subfigure}{0.32\linewidth}
        \centering
        \includegraphics[width=\linewidth]{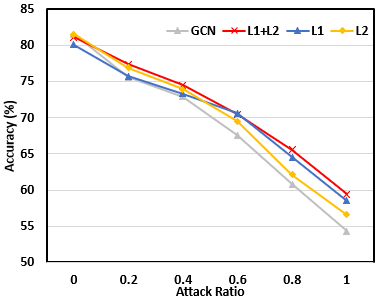}
        \caption{Benefit of Loss Designs.}
    \end{subfigure}
    \begin{subfigure}{0.32\linewidth}
        \centering
        \includegraphics[width=\linewidth]{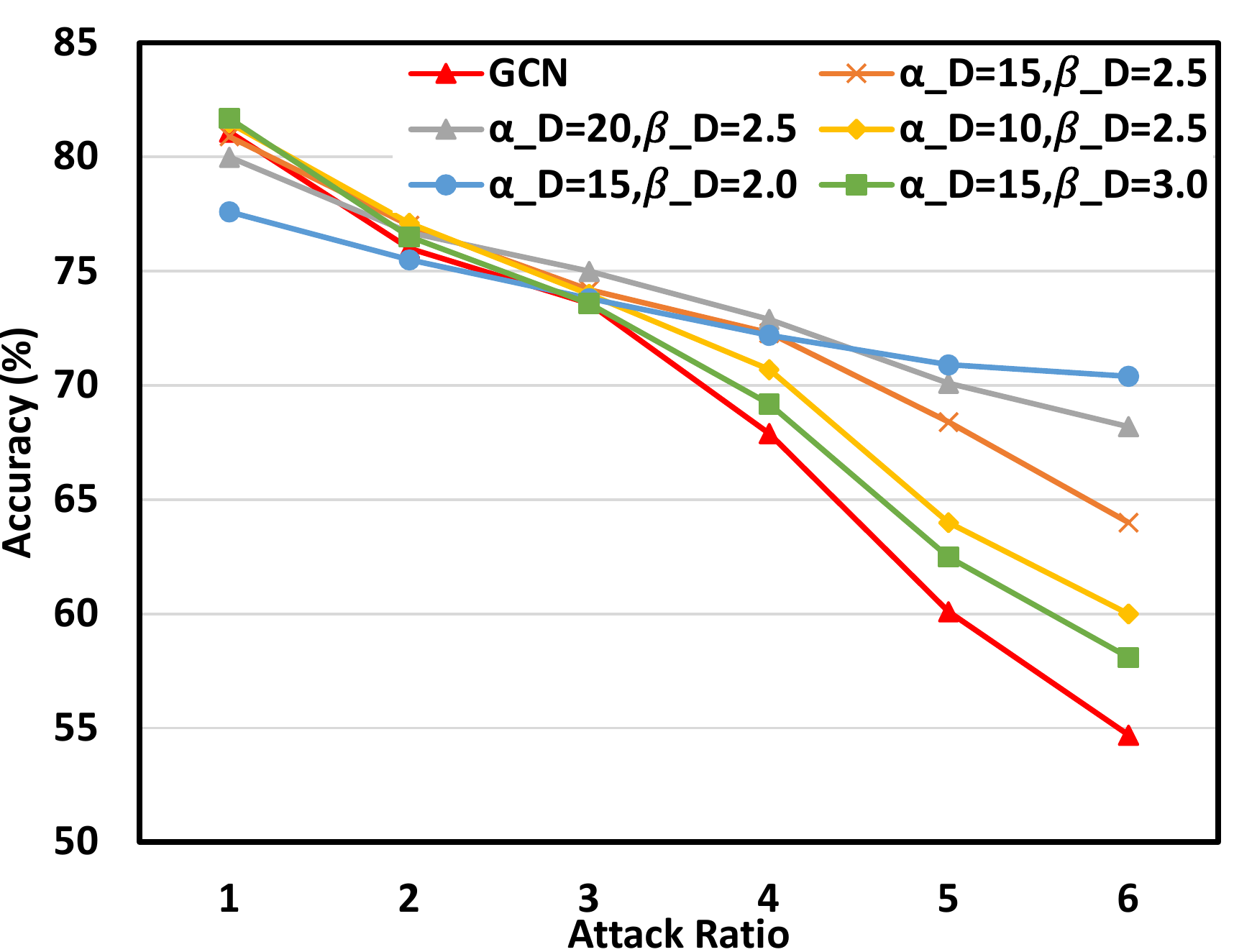}
        \caption{Impact of \textbf{Data} Hyperparameters}
    \end{subfigure}
    \begin{subfigure}{0.32\textwidth}
        \centering
        \includegraphics[width=\textwidth]{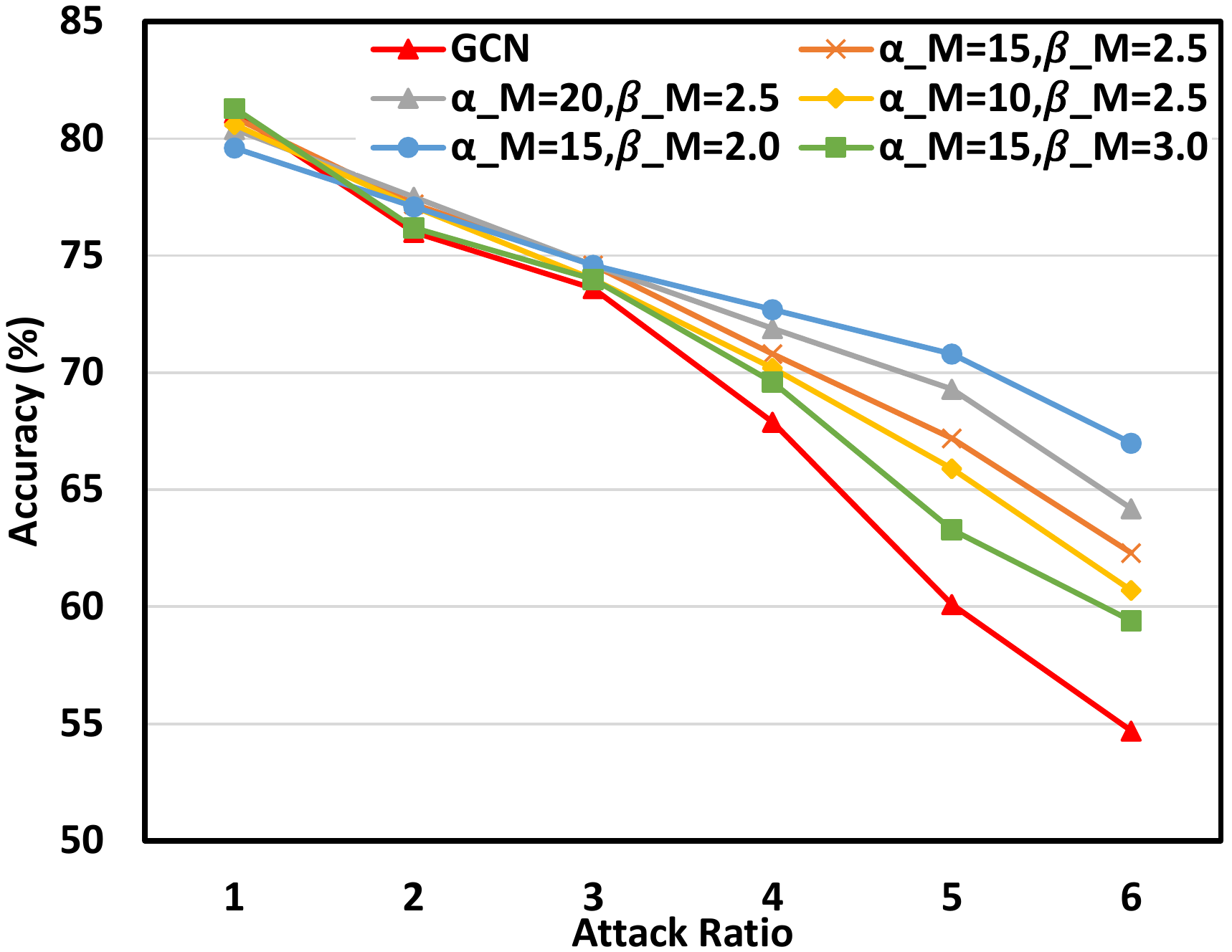}
        \caption{Impact of \textbf{Model} Hyperparameters}
    \end{subfigure}
    \vspace{-5pt}
    \caption{Loss Design Benefits and Parameter Analysis.}
    \vspace{-5pt}
    \label{fig:lossDesign_parameterAnalysis}
\end{figure*}

\begin{figure}[t]
    \begin{subfigure}{0.49\linewidth}
        \centering
        \includegraphics[width=\textwidth, height=3cm]{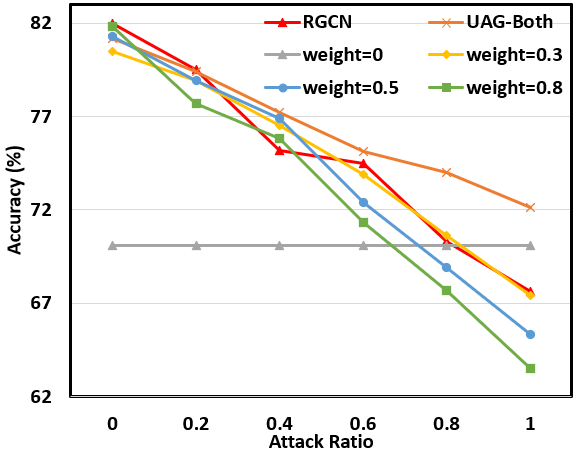}
    \end{subfigure}
    \begin{subfigure}{0.49\linewidth}
        \centering
         \includegraphics[width=\textwidth, height=2.9cm, width=3.8cm, trim=0 -0.1cm 0 0.3cm]{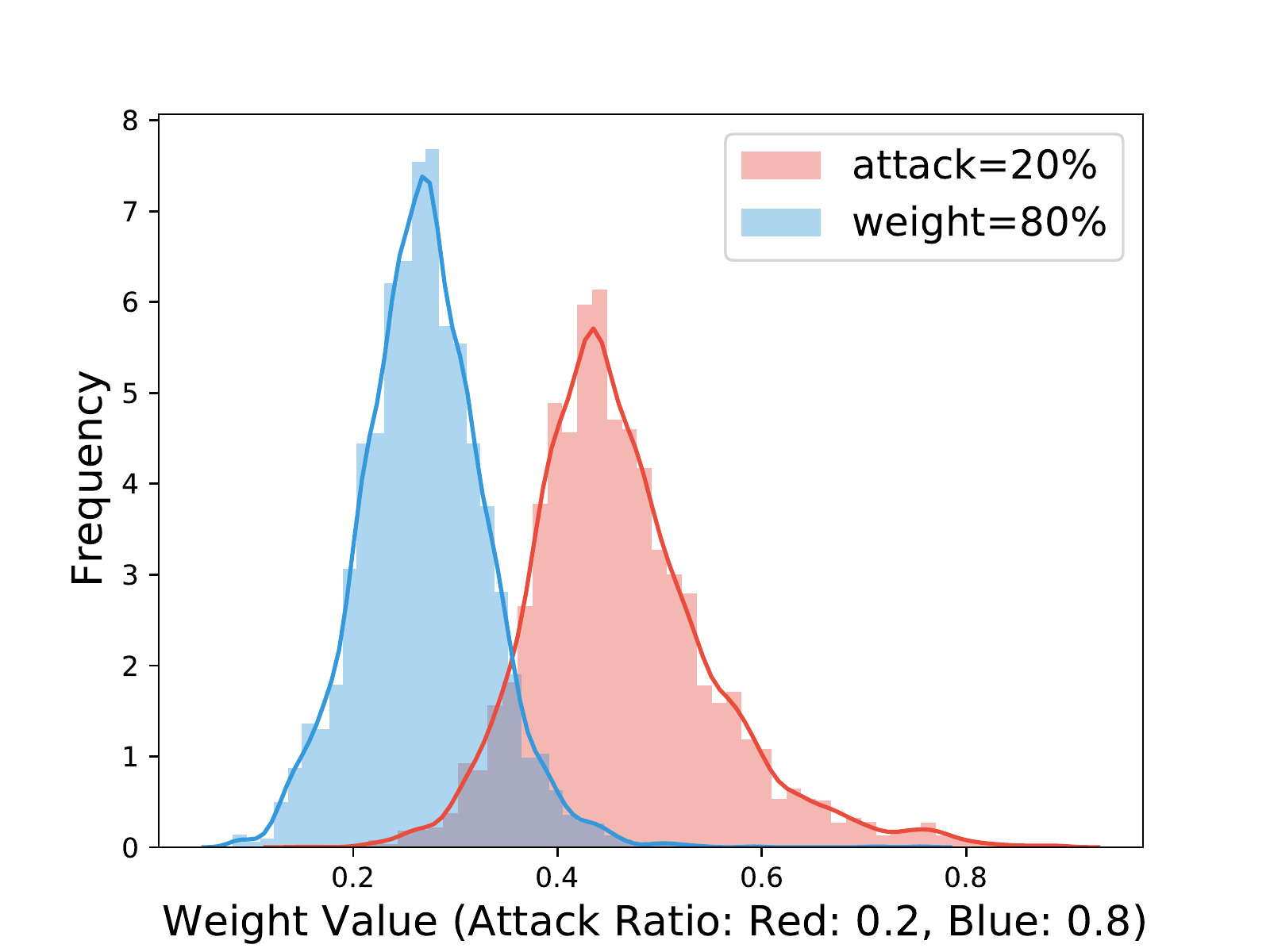}
    \end{subfigure}
    \vspace{-5pt}
    \caption{
    \textbf{Left}: Accuracy of Static Edge Weights. \textbf{Right}: Edge weight distribution under Random Attack.
    }
    \vspace{-10pt}
    \label{fig:RelationshipUncertainty_Attention_values}
\end{figure}

\section{Evaluation} \label{sec:eval}
In this section, we evaluate \Mname~on three popular datasets and compare \Mname~with three baselines to show its effectiveness.

\subsection{Experiment Environments}
\noindent \textbf{Datasets}
We select the most typical datasets (\textit{Cora}, \textit{Citeseer}, and \textit{Pubmed}) used by many GNN papers \cite{GCNConv, GINConv, SageConv} to evaluate our \Mname.
In these datasets, the node represents documents, edge refers to citations, and each node has its own associated bag-of-word features.
Table \ref{tab:dataset} summarizes the details of these datasets.
We follow the common data split by selecting 10\% nodes as the training dataset, 10\% nodes as the validation dataset, and 80\% nodes as the testing dataset. 

\vspace{3pt}
\noindent \textbf{Baselines}
\underline{\textit{Graph Convolutional Network}} (\textbf{GCN})~\cite{GCNConv} is one of the most popular GNN architectures. It has been widely adopted in node classification, graph classification, and link prediction tasks.
Besides, it is also the key backbone network for many other GNNs, such as GraphSage~\cite{SageConv}, and differentiable pooling (Diffpool)~\cite{diffpool}. \underline{\textit{Graph Attention Network}} (\textbf{GAT})~\cite{GINConv}, another typical type of GNN, aims to distinguish the graph-structure that cannot be identified by GCN. 
GAT differs from GCN in its aggregation function, which assigns different weights to different nodes during the aggregation. \underline{\textit{Robust GCN}} (\textbf{RGCN})~\cite{zhu2019robust}, leverages the Gaussian distributions for node representations to amortize the effects of adversarial attacks.

\vspace{3pt}
\noindent \textbf{Attack Methods}
\underline{\textit{Random Attack}} is a popular attack method that randomly adds fake edges into the graph dataset without considering the label of nodes.
\underline{\textit{DICE Attack}} (delete edges internally, connect externally)~\cite{DICE} exploits the node label information to increase node diversity by deleting edges between nodes with the same label and adding edges between nodes with different labels.
\underline{\textit{Mettack}} is another representative attack that adopts a meta-learning approach to reason the loss change by iteratively perturbing individual edges and features.

\vspace{3pt}
\noindent \textbf{Platforms.}
We implement \Mname~based on PyTorch Geometric \cite{PyG}. 
We evaluate \Mname~on a Dell Workstation T7910 (Ubuntu 18.04) with an Intel Xeon CPU E5-2603, 64 GB memory, and an NVIDIA 1080Ti GPU with 12GB memory.

\begin{table}[t] 
    \caption{Datasets for Evaluation.}
    \label{tab:dataset}
    \centering
    \scalebox{1}{
     \begin{tabular}{ l c c c c }
    \Xhline{2\arrayrulewidth}
    \textbf{Dataset} & \textbf{\#Vertex} & \textbf{\#Edge} & \textbf{\#Dim} & \textbf{{\#Class}}\\
    \hline
    Citeseer    & 3,327	    & 9,464	    & 3,703 & 6      \\
    Cora	    & 2,708     & 10,858	& 1,433 & 7      \\
    Pubmed	    & 19,717	& 88,676	& 500  & 3      \\
    \Xhline{2\arrayrulewidth}
    \end{tabular}
    }
    \vspace{-10pt}
\end{table}

\subsection{\textbf{Overall Performance}}
In this section, we demonstrate the effectiveness of our proposed UAG approach (\textit{UAG-both}) by comparing accuracy (under different attack methods and different attack ratio) with the original unoptimized GCN and GAT model as well as the one equipped with the state-of-the-art \textit{RGCN}~\cite{zhu2019robust} defense method.
Besides, to gain more design insights, we add two more baselines for comparison, including \textit{UAG-Data} (only considering data uncertainty) and \textit{UAG-Model} (only considering data uncertainty).

Figure~\ref{fig:Random} shows accuracy performance comparisons for \textbf{Random Attack} among different implementations on the Cora, Citeseer, and Pubmed datasets.
As we can easily tell from those three sub-plots, \Mname~method and its variants (UAG-Data, UAG-Model, and UAG-Both) consistently outperform the state-of-the-art RGCN defense approach.
The major source of such performance improvements is that \Mname~effectively captures the data uncertainty and the model uncertainty, based on which \Mname~adaptively adjusts the edge weights and the amount of information propagation between nodes.
For individual dataset settings, with the increase of the attack ratio, we see the overall trend of accuracy decreasing among these implementations. 
We also notice that on Cora and Citeseer dataset, RGCN offers notable accuracy improvement over the original GAT and GCN.
However, it is still inferior compared with our UAG approach, since we explicitly capture the uncertainties, instead of relying on a Gaussian distribution to implicitly defend adversarial attacks.
Another observation is that UAG-Both usually outperforms UAG-Data and UAG-Model under diverse datasets and attack ratios.
The reason is that the data uncertainty and the model uncertainty capture uncertainties from different sources and combining these two values usually offers a more comprehensive measurement on the prediction uncertainties.


On \textbf{DICE Attack} (Figure~\ref{fig:DICE}), besides the similar observations as the above attack setting, we have more observations. GAT and RGCN are sensitive towards DICE attack as demonstrated with significant accuracy drop with the increase of attack ratio, because DICE intentionally increases the node diversity by adding cross-community connections (\textit{i.e.}, edges between nodes with different labels).
In contrast, our UAG approach can handle this attack effectively, because it explicitly captures the node diversity as the data uncertainty.
We also observe a similar performance trend on \textbf{Mettack} (Figure~\ref{fig:Mettack}) as the previous two types of attacks and further demonstrate the advantage of our UAG approach in terms of higher accuracy under diverse attack ratios.

\subsection{\textbf{Ablation Studies}}
In this section, we conduct a set of ablation studies for in-depth analysis.

\vspace{3pt}
\noindent \textbf{Accuracy and Uncertainty.} As shown in Figure~\ref{fig:RelationshipAccuracy_Uncertainty_Diversity}(a) and (b), the increase of the model uncertainty and data uncertainty would lead to the decrease of the model accuracy.
This also strengthens our initial assumption that the relationship between uncertainties and model accuracy can be explored to defend adversarial attacks.
Figure~\ref{fig:RelationshipAccuracy_Uncertainty_Diversity}(c) exhibits the relation between data uncertainty and node diversity, showing the effectiveness of our Loss design (Eq~\ref{eq:loss1}) in learning the node diversity.

\vspace{3pt}
\noindent \textbf{Loss Design Benefits.}
Figure \ref{fig:lossDesign_parameterAnalysis}(a) validates the effectiveness of the two loss designs in the data uncertainty.
Here, L1 (Eq~\ref{eq:loss1}) forces UAG to learn the node diversity in labeled nodes, while L2 (Eq
~\ref{eq:loss2}) is an unsupervised loss that implicitly encodes the node diversity in unlabeled nodes.
We observe that the UAG with L1+L2 outperforms only L1 or L2, since it fully exploits both the labeled and unlabeled node.

\vspace{3pt}
\noindent \textbf{Parameter Analysis.}
Figure \ref{fig:lossDesign_parameterAnalysis}(b) and (c) shows the impact from different values of the data-uncertainty-related hyperparameters and the model-uncertainty-related hyperparameters on the UAG performance.
Intuitively, larger values of $\alpha_{\tau}$ and lower value of $\beta_\tau$ lead to a stronger impact from the uncertainties and higher accuracy under large attack ratios, where $\tau \in \{M,D\}$.
However, setting $\alpha_\tau$ too large or $\beta_\tau$ too small may also impose too many constraints on the information propagation.
In our experiments, we can achieve satisfying results by using $\alpha_{\tau}=15$ and $\beta_{\tau}=2.5$.

\vspace{3pt}
\noindent \textbf{Uncertainty on Attention Values.} Figure~\ref{fig:RelationshipUncertainty_Attention_values} visualizes the weight changes under different ratios of Random Attack.
We can notice that a higher random attack ratio would lead to denser weight distribution towards zero. This is because our UAG approach would try to amortize the impact of such attacks by changing weight values towards zero that can minimize the value propagation between neighboring nodes. 
Furthermore, we consider pre-assigning different weights to show the key importance of weight value for adversarial attacks. We can see that the higher the static weight value the poor the performance in maintaining model accuracy under the attack. The major reason is more ``attack impacts'' will be propagated to different nodes through node aggregation, thus, lowering the model overall performance. Our UAG, on the other hand, can adaptively determine the weight value for different nodes based on the model and data uncertainty factor, thus, largely absorbing the influence of the adversarial attack.

\vspace{-5pt}
\section{Conclusion}\label{conclusion}
In this paper, we propose \Mname, the first systematic defense solution for adversarial attacks on GNNs by considering hierarchical uncertainty in GNNs. \Mname~incorporates a Bayesian Uncertainty Technique (BUT) to explicitly capture uncertainty in GNNs and further employs an Uncertainty-aware Asymmetry Attention Technique (UAT) to fortify GNNs. Extensive experiments further demonstrate \Mname's advantages over the state-of-the-art solutions. Overall, our work paves a new way of exploring uncertainty benefits in GNN research.

\newpage

\bibliography{reference}

\begin{thebibliography}{32}
\providecommand{\natexlab}[1]{#1}
\providecommand{\url}[1]{\texttt{#1}}
\providecommand{\urlprefix}{URL }
\expandafter\ifx\csname urlstyle\endcsname\relax
  \providecommand{\doi}[1]{doi:\discretionary{}{}{}#1}\else
  \providecommand{\doi}{doi:\discretionary{}{}{}\begingroup
  \urlstyle{rm}\Url}\fi

\bibitem[{Alex~Kendall and Cipolla(2017)}]{BMVC2017_57}
Alex~Kendall, V.~B.; and Cipolla, R. 2017.
\newblock Bayesian SegNet: Model Uncertainty in Deep Convolutional
  Encoder-Decoder Architectures for Scene Understanding.
\newblock In Tae-Kyun~Kim, Stefanos~Zafeiriou, G.~B.; and Mikolajczyk, K.,
  eds., \emph{Proceedings of the British Machine Vision Conference (BMVC)},
  57.1--57.12. BMVA Press.
\newblock ISBN 1-901725-60-X.
\newblock \doi{10.5244/C.31.57}.
\newblock \urlprefix\url{https://dx.doi.org/10.5244/C.31.57}.

\bibitem[{Bojchevski and Günnemann(2018)}]{bojchevski2018deep}
Bojchevski, A.; and Günnemann, S. 2018.
\newblock Deep Gaussian Embedding of Graphs: Unsupervised Inductive Learning
  via Ranking.
\newblock In \emph{International Conference on Learning Representations}.
\newblock \urlprefix\url{https://openreview.net/forum?id=r1ZdKJ-0W}.

\bibitem[{Chen, Li, and Huang(2005)}]{chen2005link}
Chen, H.; Li, X.; and Huang, Z. 2005.
\newblock Link prediction approach to collaborative filtering.
\newblock In \emph{Proceedings of the 5th ACM/IEEE-CS Joint Conference on
  Digital Libraries (JCDL)}, 141--142. IEEE.

\bibitem[{Cheng et~al.(2018)Cheng, Gong, Chang, Shi, Hauptmann, and
  Zheng}]{cheng2018deep}
Cheng, D.; Gong, Y.; Chang, X.; Shi, W.; Hauptmann, A.; and Zheng, N. 2018.
\newblock Deep feature learning via structured graph Laplacian embedding for
  person re-identification.
\newblock \emph{Pattern Recognition} 82: 94--104.

\bibitem[{Dai et~al.(2018)Dai, Li, Tian, Huang, Wang, Zhu, and
  Song}]{dai2018adversarial}
Dai, H.; Li, H.; Tian, T.; Huang, X.; Wang, L.; Zhu, J.; and Song, L. 2018.
\newblock Adversarial attack on graph structured data.
\newblock \emph{arXiv preprint arXiv:1806.02371} .

\bibitem[{Duran and Niepert(2017)}]{duran2017learning}
Duran, A.~G.; and Niepert, M. 2017.
\newblock Learning graph representations with embedding propagation.
\newblock In \emph{Advances in neural information processing systems (NIPS)},
  5119--5130.

\bibitem[{Fey and Lenssen(2019)}]{PyG}
Fey, M.; and Lenssen, J.~E. 2019.
\newblock Fast Graph Representation Learning with {PyTorch Geometric}.
\newblock In \emph{ICLR Workshop on Representation Learning on Graphs and
  Manifolds (ICLR)}.

\bibitem[{Gal and Ghahramani(2016{\natexlab{a}})}]{dropoutBayesian}
Gal, Y.; and Ghahramani, Z. 2016{\natexlab{a}}.
\newblock Dropout as a Bayesian Approximation: Representing Model Uncertainty
  in Deep Learning.
\newblock In \emph{Proceedings of the 33rd International Conference on
  International Conference on Machine Learning - Volume 48}, ICML’16,
  1050–1059. JMLR.org.

\bibitem[{Gal and Ghahramani(2016{\natexlab{b}})}]{dropoutRNN}
Gal, Y.; and Ghahramani, Z. 2016{\natexlab{b}}.
\newblock A Theoretically Grounded Application of Dropout in Recurrent Neural
  Networks.
\newblock In \emph{Proceedings of the 30th International Conference on Neural
  Information Processing Systems}, NIPS’16, 1027–1035. Red Hook, NY, USA:
  Curran Associates Inc.
\newblock ISBN 9781510838819.

\bibitem[{Gibert, Valveny, and Bunke(2012)}]{gibert2012graph}
Gibert, J.; Valveny, E.; and Bunke, H. 2012.
\newblock Graph embedding in vector spaces by node attribute statistics.
\newblock \emph{Pattern Recognition} 45(9): 3072--3083.

\bibitem[{Grover and Leskovec(2016)}]{grover2016node2vec}
Grover, A.; and Leskovec, J. 2016.
\newblock node2vec: Scalable feature learning for networks.
\newblock In \emph{Proceedings of the 22nd ACM international conference on
  Knowledge discovery and data mining (SIGKDD)}, 855--864.

\bibitem[{Hamilton, Ying, and Leskovec(2017)}]{SageConv}
Hamilton, W.; Ying, Z.; and Leskovec, J. 2017.
\newblock Inductive representation learning on large graphs.
\newblock In \emph{Advances in neural information processing systems (NIPS)},
  1024--1034.

\bibitem[{Hasanzadeh et~al.(2020)Hasanzadeh, Hajiramezanali, Boluki, Zhou,
  Duffield, Narayanan, and Qian}]{hasanzadeh2020bayesian}
Hasanzadeh, A.; Hajiramezanali, E.; Boluki, S.; Zhou, M.; Duffield, N.;
  Narayanan, K.; and Qian, X. 2020.
\newblock Bayesian Graph Neural Networks with Adaptive Connection Sampling.
\newblock In \emph{ICML 2020: International Conference on Machine Learning}.
\newblock \urlprefix\url{https://arxiv.org/abs/2006.04064}.

\bibitem[{Kaspar and Horst(2010)}]{kaspar2010graph}
Kaspar, R.; and Horst, B. 2010.
\newblock \emph{Graph classification and clustering based on vector space
  embedding}, volume~77.
\newblock World Scientific.

\bibitem[{Kendall and Gal(2017)}]{kendall2017uncertainties}
Kendall, A.; and Gal, Y. 2017.
\newblock What uncertainties do we need in bayesian deep learning for computer
  vision?
\newblock In \emph{Advances in neural information processing systems
  (NeurIPS)}, 5574--5584.

\bibitem[{Kipf and Welling(2017)}]{GCNConv}
Kipf, T.~N.; and Welling, M. 2017.
\newblock Semi-supervised classification with graph convolutional networks.
\newblock \emph{International Conference on Learning Representations (ICLR)} .

\bibitem[{Kunegis and Lommatzsch(2009)}]{kunegis2009learning}
Kunegis, J.; and Lommatzsch, A. 2009.
\newblock Learning spectral graph transformations for link prediction.
\newblock In \emph{Proceedings of the 26th Annual International Conference on
  Machine Learning (ICML)}, 561--568.

\bibitem[{Luo et~al.(2009)Luo, Ding, Huang, and Li}]{luo2009non}
Luo, D.; Ding, C.; Huang, H.; and Li, T. 2009.
\newblock Non-negative laplacian embedding.
\newblock In \emph{2009 Ninth IEEE International Conference on Data Mining
  (ICDM)}, 337--346. IEEE.

\bibitem[{Luo et~al.(2011)Luo, Nie, Huang, and Ding}]{luo2011cauchy}
Luo, D.; Nie, F.; Huang, H.; and Ding, C.~H. 2011.
\newblock Cauchy graph embedding.
\newblock In \emph{Proceedings of the 28th International Conference on Machine
  Learning (ICML)}, 553--560.

\bibitem[{Perozzi, Al-Rfou, and Skiena(2014)}]{deepWalk}
Perozzi, B.; Al-Rfou, R.; and Skiena, S. 2014.
\newblock DeepWalk: Online Learning of Social Representations.
\newblock In \emph{Proceedings of the 20th ACM International Conference on
  Knowledge Discovery and Data Mining (SIGKDD)}, 701–710. New York, NY, USA:
  Association for Computing Machinery.
\newblock ISBN 9781450329569.
\newblock \doi{10.1145/2623330.2623732}.
\newblock \urlprefix\url{https://doi.org/10.1145/2623330.2623732}.

\bibitem[{Tylenda, Angelova, and Bedathur(2009)}]{tylenda2009towards}
Tylenda, T.; Angelova, R.; and Bedathur, S. 2009.
\newblock Towards time-aware link prediction in evolving social networks.
\newblock In \emph{Proceedings of the 3rd workshop on social network mining and
  analysis}, 1--10.

\bibitem[{Veličković et~al.(2018)Veličković, Cucurull, Casanova, Romero,
  Liò, and Bengio}]{GATConv}
Veličković, P.; Cucurull, G.; Casanova, A.; Romero, A.; Liò, P.; and Bengio,
  Y. 2018.
\newblock Graph Attention Networks.
\newblock In \emph{International Conference on Learning Representations
  (ICLR)}.
\newblock \urlprefix\url{https://openreview.net/forum?id=rJXMpikCZ}.

\bibitem[{Waniek et~al.(2018)Waniek, Michalak, Wooldridge, and Rahwan}]{DICE}
Waniek, M.; Michalak, T.~P.; Wooldridge, M.~J.; and Rahwan, T. 2018.
\newblock Hiding individuals and communities in a social network.
\newblock \emph{Nature Human Behaviour} 2: 139.

\bibitem[{Xiao and Wang(2019)}]{xiao2019quantifying}
Xiao, Y.; and Wang, W.~Y. 2019.
\newblock Quantifying uncertainties in natural language processing tasks.
\newblock In \emph{Proceedings of the AAAI Conference on Artificial
  Intelligence}, volume~33, 7322--7329.

\bibitem[{Xu et~al.(2019{\natexlab{a}})Xu, Chen, Liu, Chen, Weng, Hong, and
  Lin}]{xu2019topology}
Xu, K.; Chen, H.; Liu, S.; Chen, P.-Y.; Weng, T.-W.; Hong, M.; and Lin, X.
  2019{\natexlab{a}}.
\newblock Topology attack and defense for graph neural networks: An
  optimization perspective.
\newblock \emph{arXiv preprint arXiv:1906.04214} .

\bibitem[{Xu et~al.(2019{\natexlab{b}})Xu, Hu, Leskovec, and Jegelka}]{GINConv}
Xu, K.; Hu, W.; Leskovec, J.; and Jegelka, S. 2019{\natexlab{b}}.
\newblock How Powerful are Graph Neural Networks?
\newblock In \emph{International Conference on Learning Representations
  (ICLR)}.
\newblock \urlprefix\url{https://openreview.net/forum?id=ryGs6iA5Km}.

\bibitem[{Ying et~al.(2018)Ying, You, Morris, Ren, Hamilton, and
  Leskovec}]{diffpool}
Ying, R.; You, J.; Morris, C.; Ren, X.; Hamilton, W.~L.; and Leskovec, J. 2018.
\newblock Hierarchical Graph Representation Learning with Differentiable
  Pooling.
\newblock In \emph{Proceedings of the 32nd International Conference on Neural
  Information Processing Systems (NIPS)}, 4805–4815. Red Hook, NY, USA.

\bibitem[{Zhang et~al.(2019)Zhang, Pal, Coates, and Ustebay}]{BGCN}
Zhang, Y.; Pal, S.; Coates, M.; and Ustebay, D. 2019.
\newblock Bayesian Graph Convolutional Neural Networks for Semi-Supervised
  Classification.
\newblock \emph{Proceedings of the AAAI Conference on Artificial Intelligence}
  33: 5829--5836.
\newblock \doi{10.1609/aaai.v33i01.33015829}.

\bibitem[{Zhu et~al.(2019)Zhu, Zhang, Cui, and Zhu}]{zhu2019robust}
Zhu, D.; Zhang, Z.; Cui, P.; and Zhu, W. 2019.
\newblock Robust graph convolutional networks against adversarial attacks.
\newblock In \emph{Proceedings of the 25th ACM SIGKDD International Conference
  on Knowledge Discovery \& Data Mining}, 1399--1407.

\bibitem[{{Zhu} and {Laptev}(2017)}]{uberUncertainty}
{Zhu}, L.; and {Laptev}, N. 2017.
\newblock Deep and Confident Prediction for Time Series at Uber.
\newblock In \emph{2017 IEEE International Conference on Data Mining Workshops
  (ICDMW)}, 103--110.

\bibitem[{Z{\"u}gner, Akbarnejad, and
  G{\"u}nnemann(2018)}]{zugner2018adversarial}
Z{\"u}gner, D.; Akbarnejad, A.; and G{\"u}nnemann, S. 2018.
\newblock Adversarial attacks on neural networks for graph data.
\newblock In \emph{Proceedings of the 24th ACM SIGKDD International Conference
  on Knowledge Discovery \& Data Mining}, 2847--2856.

\bibitem[{Z{\"u}gner and G{\"u}nnemann(2019)}]{zugner2019adversarial}
Z{\"u}gner, D.; and G{\"u}nnemann, S. 2019.
\newblock Adversarial attacks on graph neural networks via meta learning.
\newblock \emph{arXiv preprint arXiv:1902.08412} .

\end{thebibliography}

\end{document}